\documentclass{bmvc2k}

\usepackage{graphicx} 
\usepackage{caption}
\usepackage{subcaption}
\usepackage{pifont}
\usepackage{bbding}
\usepackage{colortbl}
\usepackage{booktabs}
\usepackage{amssymb}
\usepackage{xcolor} 
\usepackage{hyperref}  % 允许添加链接
\hypersetup{
    urlcolor=magenta,      % 设置 URL 链接颜色
}
\addauthor{Zhi Cai}{caizhi97@buaa.edu.cn}{1,2}
\addauthor{Songtao Liu}{liusongtao@megvii.com}{3}
\addauthor{Guodong Wang}{wangguodong@buaa.edu.cn}{1,2}
\addauthor{Zheng Ge}{gezheng@megvii.com}{3}
\addauthor{Zeming Li}{lizeming@megvii.com}{3}
\addauthor{Xiangyu Zhang}{zhangxiangyu@megvii.com}{3}
\addauthor{Di Huang}{dhuang@buaa.edu.cn}{1,2,$^{\dag}$}

\addinstitution{
 SKLCCSE, Beihang University\\
 Beijing, China
}
\addinstitution{
 IRIP Lab, SCSE, Beihang University\\
 Beijing, China
}
\addinstitution{
 Megvii Inc.\\
}
\runninghead{Z. Cai et al.}{Align-DETR: Enhancing DETR with aligned loss}

\title{Align-DETR: Enhancing End-to-end Object Detection with Aligned Loss}

\begin{document}

\newcommand{\hy}{\hat{y}}
\newcommand{\hb}{\hat{b}}
\newcommand{\hp}{\hat{p}}
\newcommand{\ty}{\tilde{y}}
\newcommand{\dabplus}{$\text{DAB-DETR}^+$}
\newcommand{\methodname}{Align-DETR~}
\newcommand{\AP}[1]{ $ \text{AP}_{#1}$}
\newcommand{\bsigma}{\mathfrak{S}}
\newcommand{\loss}[1]{{\cal L}(#1)}
\newcommand{\closs}[2]{{\cal L}_{\rm class}(#1, #2)}
\newcommand{\bloss}[1]{{\cal L}_{\rm box}(#1)}
\newcommand{\iouloss}[1]{{\cal L}_{\rm iou}(#1)}
\newcommand{\diceloss}[1]{{\cal L}_{\rm DICE}(#1)}
\newcommand{\hloss}[1]{{\cal L}_{\rm H}(#1)}
\newcommand{\green}[1]{\textcolor{green}{#1}}
\newcommand{\red}[1]{\textcolor{red}{#1}}
\newcommand{\blue}[1]{\textcolor{blue}{#1}}

%%%%%%%%% TITLE

% \and
% }10

\maketitle

%%%%%%%%% ABSTRACT
\begin{abstract}
DETR has set up a simple end-to-end pipeline for object detection by formulating this task as a set prediction problem, showing promising potential. 
Despite its notable advancements, this paper identifies two key forms of misalignment within the model: classification-regression misalignment and cross-layer target misalignment. 
Both issues impede DETR's convergence and degrade its overall performance.
To  tackles both issues simultaneously, we introduce a novel loss function, termed as  Align Loss, designed to resolve the discrepancy between the two tasks.
Align Loss guides the optimization of DETR through a joint quality metric, strengthening the connection between classification and regression. 
Furthermore, it incorporates an exponential down-weighting term to facilitate a smooth transition from positive to negative samples. 
Align-DETR also employs many-to-one matching for supervision of intermediate layers, akin to the design of  $\mathcal{H}$-DETR , which enhances robustness against instability.
We conducted extensive experiments, yielding highly competitive results. Notably, our method achieves a $49.3\%~(+0.6)$ AP on the $\mathcal{H}$-DETR baseline with the ResNet-50 backbone. It also sets a new state-of-the-art performance, reaching $50.5\%$ AP in the 1$\times$ setting and $51.7\%$ AP in the 2$\times$ setting, surpassing several strong competitors. Our code is  available at \href{https://github.com/FelixCaae/AlignDETR}{https://github.com/FelixCaae/AlignDETR}.

\end{abstract}

\section{Introduction}

%First 
%First 
Recently, transformer-based methods have garnered significant attention in the object detection community, largely due to the introduction of the DETR paradigm by \cite{detr}. Unlike previous CNN-based detectors~\cite{faster,atss,autoassign,focalloss},
DETR approaches object detection as a set prediction problem, utilizing learnable queries to represent each object in one-to-one correspondence.  Such unique correspondence derives from bipartite graph matching by means of label assignment during training. It bypasses hand-crafted components such as non-maximum suppression (NMS) and anchor generation. With this simple and extensible pipeline, DETR shows great potential in a wide variety of areas, including 2D segmentation~\cite{maskformer,dong2021solq,maskdino}, 3D detection ~\cite{detr3d,3detr,petr}, in addition to 2D detection ~\cite{sparsercnn, deformabledetr, adamixer, dino,liu2023stable, hu2024dac}.

\begin{figure}[t]
\centering
\includegraphics[width=0.4\columnwidth]{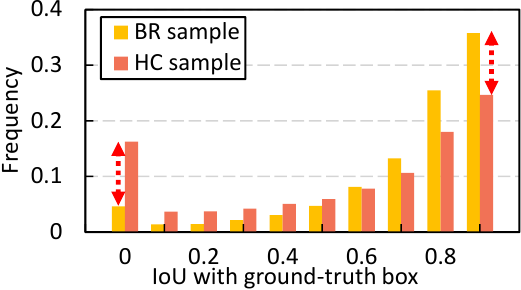}
\includegraphics[width=0.35\columnwidth]{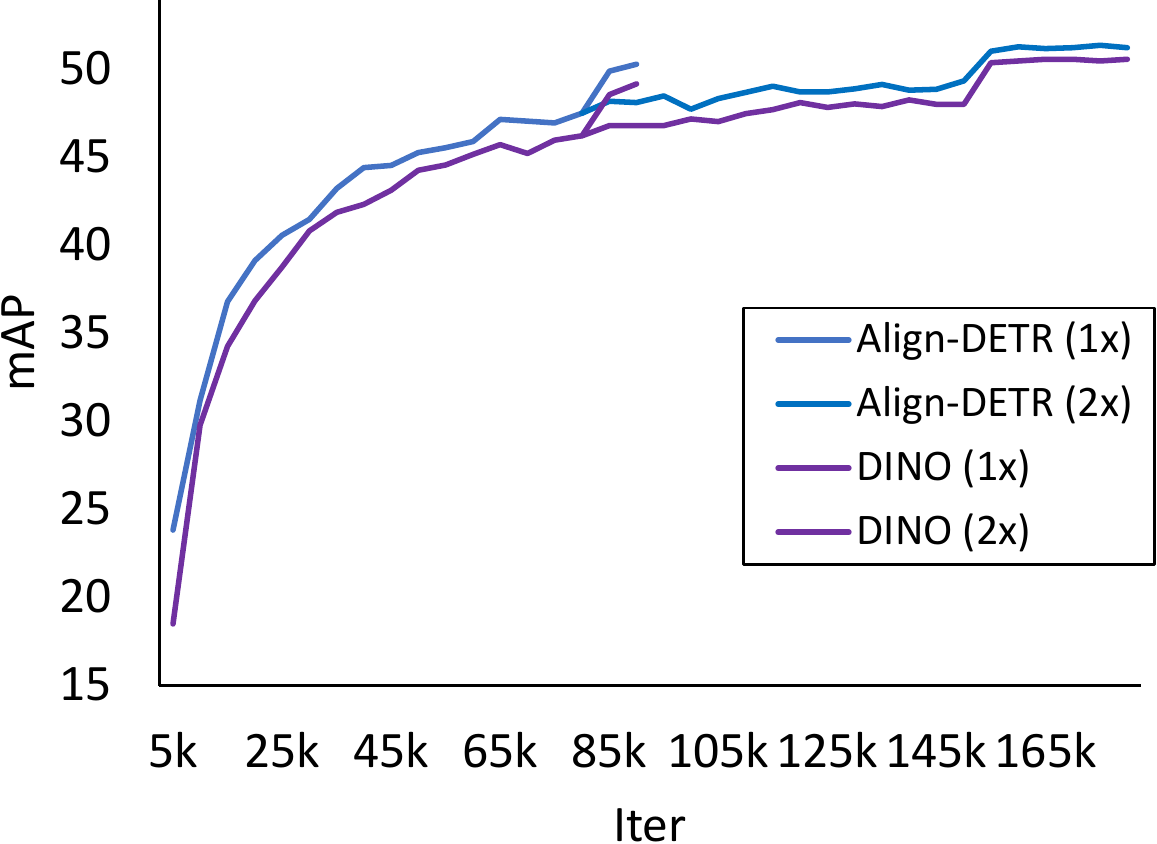}
\caption{
    \textbf{Left:} Intersection over Union (IoU) distribution of two types of samples. There is a notable gap between best regressed samples (oracle) and the high confident samples, indicating a discrepancy between these two tasks. \textbf{Right:} The convergence curve of Align-DETR and DINO where Align-DETR converges faster significantly.}
% The huge gap between these two kinds of samples indicates that there is huge space of improvement.}
 \label{fig:fig1}
\end{figure}

% which explicitly uses intermediate predictions as the next stages' priors.
During the past few years, the successors have advanced DETR in many ways. For instance, some methods attempt to incorporate local operators, such as ROI pooling~\cite{sparsercnn} or deformable attention~\cite{deformabledetr,adamixer}, to increase the convergence speed and reduce the computational cost; some methods indicate that those learnable queries can be improved through extra physical embeddings~\cite{conddetr, anchordetr, dabdetr}; and some methods~\cite{dndetr, hdetr, chen2023group, dino} notice the defect of one-to-one matching and introduce more positive samples by adding training-only queries. Box refinement~\cite{deformabledetr,sparsercnn,dino} is another helpful technique, which explicitly takes previous predictions as priors at the next stages.

% \begin{figure*}[t]
%     \centering
%     \includegraphics[width=1.0\linewidth]{figures/soto_matching.pdf}
%     \caption{A simplified illustration of the difference between SOTO matching and other matching methods. }
%     \label{fig:concept}
% \end{figure*}

Despite the recent progress in DETR-based detectors~\cite{detr,d2etr,deformabledetr,adamixer,dino,dabdetr,hu2024dac}, the misalignment problem of DETR has received insufficient attention. There are two key aspects to this misalignment issue in recent DETR-like methods. 
Firstly, there exists a misalignment between classification confidence and localization precision, stemming from inconsistent loss design. This discrepancy is highlighted through an analysis conducted on the output of a prominent end-to-end detector, DINO~\cite{dino}, revealing a significant dissonance between high-confidence samples (HC samples) and best-regressed samples (BR samples), as depicted in Fig.~\ref{fig:fig1} (Left). Such a discrepancy significantly impacts model performance, particularly in ranking-based metrics such as mean average precision (mAP).
Secondly, there is a misalignment in training targets across layers. This arises from the dynamic matching design of DETR~\cite{dndetr,dino,liu2023stable}, wherein samples are assigned different targets in different layers, leading to confusion within the optimizer, as highlighted by Stable-DINO~\cite{liu2023stable}. These misalignment issues impede the convergence of DETR and hinder it from realizing its full potential as shown in Fig.~\ref{fig:fig1} (Right).
% Stable-DINO proposes a position-guided loss for DETR, however, it is designed for one-to-one matching which is inefficient compared to many-to-one matching. 

% 

% This problem refers to the inconsistency of a output predictions between the classification confidence and localization precision, \emph{e.g.} a highly confident prediction with a relatively low intersection-over-union (IoU) score or vice versa~\cite{iounet,gfl}. Ideally, the predictions with the highest classification scores (HC samples) also have the best-regressed bounding boxes (BR samples); however, it does not hold when the misalignment problem occurs, creating the risk of missing BR predictions and thus deteriorating the detection performance~\cite{iounet,gfl}. 

The current solutions to the misalignment problem in DETR-like methods typically address either the first misalignment issue~\cite{iounet,vfl,gfl} or the second~\cite{dino,liu2023stable}. To tackle both simultaneously, we introduce a novel approach called Align-DETR. 
It makes use of the standard focal loss~\cite{focalloss} with an IoU-aware target on foreground samples, which we term the Align Loss. To overcome the first misalignment problem,  Align Loss dynamically adjusts the target for foreground samples according to their classification confidence and regression precision thus they are aligned during optimization~\cite{liu2023stable}. For the second problem, Align-DETR enlarges the range of positive samples by adopting a mixed-matching strategy. This approach allows multiple candidates to be considered for each ground truth. Subsequently, to mitigate conflicts arising from this expanded range of positive samples, the targets of the additional positive samples are smoothed using an exponential weight decay. By incorporating these mechanisms, Align-DETR aims to effectively address both misalignment issues encountered in DETR-based detectors.

Overall, \methodname~offers a straightforward yet effective solution to the misalignment problem, enhancing DETR with aligned training targets. Equipped with a ResNet-50~\cite{he2016deep} backbone and a $\mathcal{H}$-DETR~\cite{hdetr} baseline, our method achieves +0.6\% AP gain. We also combine it with the strong baseline DINO~\cite{dino} and establish a new state-of-the-art performance with $50.5\%$ AP in the 1$\times $ and $51.7\%$ AP in 2$\times$  setting on the COCO~\cite{coco} validation set.

\section{Related Work}
% \subsection{.}
% Modern object detectors\cite{detr,faster} usually produce multiple predictions for one instance and this requires a selection process to remove duplicates and find the optimal prediction. Traditionall, this process is set as an individual s
%The success of convolution neural networks (CNN) \cite{he2016deep,vgg} paves the way of modern object detection.  The classical CNN-based object detectors  can be roughly split into two categories, two-stage detectors\cite{faster,mask} and one-stage detectors. Two-stage object detector takes an extra step to propose proposals and one-stage object detector directly predict on the dense feature map. Anchor is widely used in modern one-stage and two-stage object detectors as initial guess of where the object locates.  Nevertheless, designing optimal anchors requires much efforts. Another widely used component, the non-maximum supression (NMS).

% \begin{figure}[ht]
%     \centering
%     \includegraphics[width=1\linewidth]{iccv2023AuthorKit/figures/fig2.pdf}
%     \caption{Comparison of the solutions to the misalignment problem.  In CNN-based object detectors, the main efforts are made to guide the NMS~\cite{iounet} better selection. While ours IA-BCE focuses on improving the model's relation modeling ability.}
%     \label{fig:fig_2}
% \end{figure}
\subsection{Label Assignment in Object Detection}

As the CNN-based object detectors develop from the anchor-based framework to the anchor-free one, many works realize the importance of label assignment (which is previously hidden by anchor and IoU matching) during training. Some works ~\cite{paa,ge2021ota,atss} identify positive samples by measuring their dynamic prediction quality for each object.
%In fact, this idea also appears in DETR and its variants where the difference is made between the many-to-one and one-to-one manner.
Others~\cite{autoassign, gfl, dw, musu} learn the assignment in a soft way and achieve better alignment on prediction quality by incorporating IoU~\cite{gfl, vfl} or a combination of IoU and confidence~\cite{autoassign,dw,musu}. 

The misalignment problem in object detection has been addressed by various traditional solutions, such as incorporating an additional IoU branch to fine-tune the confidence scores~\cite{iounet} or integrating the IoU prediction branch into classification losses~\cite{gfl}. 
In contrast, the misalignment problem in
DETR is under-explored, despite it sharing some ideas from CNN-based detectors. However, the optimization target between many-to-one and one-to-one is the key difference as we will illustrate in the next section.
\subsection{End-to-end Object Detection}

The pursuit of end-to-end object detection or segmentation dates back to several early efforts \cite{ren2017end, sal2017recur, stewart2016end}. They rely on recurrent neural network (RNN)\cite{sal2017recur} to remove duplicates or adopt complex subnets \cite{ren2017end,stewart2016end} to replace NMS. Different from them, DETR~\cite {detr} has established a set-prediction framework based on the transformer~\cite{vaswani2017}. Compared to previous work, DETR is rather simpler but still suffers from the downside of slow convergence with a number of subsequent DETR variants~\cite{sparsercnn,adamixer,smca,dabdetr,dndetr,dino} working on this issue. Some methods make improvements on the cross-attention in decoders~\cite{smca,conddetr}. Deformable DETR~\cite{deformabledetr} presents a deformable-attention module that only scans a small set of points near the reference point, while AdaMixer\cite{adamixer} further extends the 2D offset to 3D for better multi-scale feature fusion.
% Alternatively, others consider introducing prior knowledge into queries\cite{conddetr,smca,anchordetr,dabdetr}.

% For example, Conditional-DETR~\cite{conddetr} incorporates reference coordinates into the position embedding and SMCA~\cite{smca} applies a modulated Gaussian bias in the cross-attention module. 

Besides, some works also pay attention to improve the inference efficiency of DETR~\cite{pnpdetr,efficientdetr,focusdetr,rtdetr,litedetr}.
% PnP-DETR~\cite{pnpdetr} m proposes a dynamic-net based method to prune the background tokens, making a pioneering effort in this direction.  
Efficient DETR~\cite{efficientdetr} advocates for a 1-layer-only  decoder structure to largely reduce the computation burden by initializing the query precisely. Notably, RT-DETR \cite{rtdetr} represents a significant advancement by enabling real-time inference for DETR, surpassing the performance of other rapid detectors such as the YOLO series series~\cite{ge2021yolox}.
 
The optimization of DETR also attracts the attention of many researchers~\cite{dndetr,dino,liu2023stable}. Specifically, DN-DETR~\cite{dndetr} relies on a denoising mechanism to stabilize the training, which is further refined by DINO~\cite{dino} through introducing a contrastive denoising mechanism.  Additionally, Stable-DINO~\cite{liu2023stable} introduces a position-guided loss that mitigates the instability incurred by a standard loss, \emph{i.e.} focal loss\cite{focalloss}. 
Meanwhile, a few recent studies have noticed limitations of one-to-one matching and have proposed many-to-one assigning strategies to ameliorate DETR regarding training efficiency. Group-DETR~\cite{chen2023group}  and  $\mathcal{H}$-DETR~\cite{hdetr} accelerate the training process with multiple groups of samples and ground truths.
DAC-DETR\cite{hu2024dac}  proposes a decoupled training strategy that focuses on the learning of cross-attention layers with many-to-one matching.

% $\mathcal{H}$-DETR~\cite{hdetr} introduces a hybrid branch mechanism to increase the training samples without ruining the end-to-end property.
% Thanks to their efforts, the DETR pipeline benefits from efficient training but at the cost of complexity and computation burden. For example, Group-DETR~\cite{gdetr} and Hybrid-DETR~\cite{hdetr} use 3300  (11 groups) and 1800 queries (5x in extra branch), respectively. In contrast, our proposed strategy does not introduce more queries and keeps the pipeline training efficient.

Despite the strides made, it is evident that many contemporary approaches~\cite{dino,dndetr,hu2024dac,chen2023group,hdetr} either overlook the misalignment issue highlighted earlier or offer only partial remedies~\cite{liu2023stable}. In contrast to these approaches, our work offers a comprehensive and unified solution to address this challenge consistently.

\section{Method}

% \blue{As for DETR and its variants, they have  some key differences compared with CNN-based detectors.}

% \blue{\textbf{Sparsity}. DETR uses queries to interact with features, and the number of queries is small, \emph{e.g.} 100 or 300, which is much less  compared to the number of  anchors  used in CNN-based object detectors. 
% }

% \blue{\textbf{End-to-end.}  Post-processing steps such as NMS are widely adopted in classical object detectors to remove duplicates. In contrast, DETR is an end-to-end object detector by the roots.
% }

% \blue{\textbf{Dynamic matching.} Queries in DETR can attend to the whole image and they are matched with GT by a global dynamic matching process. While CNN-based object detectors widely adopt local prior such as IoU as an important indicator.}

% On the other side, DETR and its variants generally adopt one-to-one label assignment, which proves essential to the end-to-end pipeline, also evidenced by some exploring investigations in CNN-based models~\cite{e2efcn, nmsfree}. However,  it assigns only one positive sample for each GT and hence slows down convergence. 

% remember to cite DETA

%In this section we revisit some important previous works first and then we demonstrate our method.
\subsection{Preliminaries}
\textbf{DETR}. The original DETR~\cite{detr} framework consists of three main components: a CNN-backbone, an encoder-decoder transformer~\cite{vaswani2017}, and a prediction head. The backbone processes the input image first, and the resulting feature is flattened into a series of tokens $X=\{x_1,x_2,...,x_m\}$. Then the transformer extracts information from $X$ with a group of learnable queries $Q=\{q_1,q_2,...,q_n\}$ as containers. At last, the updated queries are transformed into predictions $P=\{p_1,p_2,...,p_n\}$ through the prediction head. In most cases, $m$ is much less than $n$, making DETR a sparse object detection pipeline. 

The focal loss~\cite{focalloss} is adopted by DETR in classification optimization to help focus on important samples. Given a binary label $y\in\{0,1\}$ and a logit $p\in[0,1]$, it is defined as:

\begin{equation}
\label{eq:focal_loss}
    {\cal L}_{focal} = -y\cdot (1-p)^\gamma \cdot logp - (1-y) \cdot p^\gamma \cdot log(1-p),
\end{equation}
where $\gamma$ is the hyper-parameter to control the degree of weight decay.

% \textbf{Hybrid Layer Matching}.
DETR adopts the one-to-one label assignment on all layers to help eliminate redundant predictions. However, this strategy is inefficient compared to many-to-one label assignment used in CNN-based detectors\cite{faster,atss,autoassign}. To overcome this issue, $\mathcal{H}$-DETR\cite{hdetr} proposes a hybrid layer matching strategy that applies many-to-one matching on some shallow layers and one-to-one matching on deep layers.  Hybrid matching ensures DETR's final outputs are unique while it allows more efficient training on intermediate layers.

\subsection{Motivation and Framework} 
The motivation for the proposed \methodname comes from the hypothesis that a consistent and aligned optimization target can benefit the training of object detectors like DETR~\cite{deformabledetr,dino,dndetr,liu2023stable}. 
There are two concerns for the current optimization method: (i) the alignment between classification and regression is essential for the optimization of DETR, which is not considered in current design and (ii) the matching mechanism of DETR is unstable across layers.
To mitigate the concerns, we propose a unified solution,  namely Align-DETR.

\begin{figure*}[ht]
\centering
\includegraphics[width=0.9\linewidth]{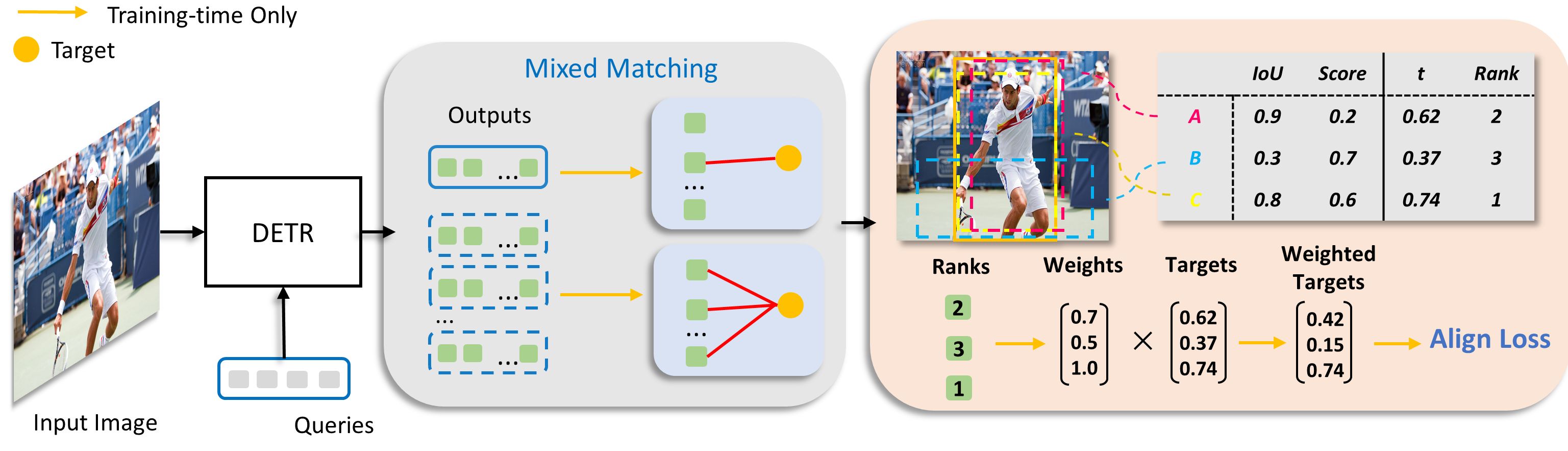}
\caption{The architecture overview of the proposed approach \methodname. Align-DETR adopts many-to-one matching  where each GT is assigned multiple queries. These queries are sorted according to their  quality . Then, we compute an alignment score for each query according to their rank, classification confidence and IoU with the GT. The alignment score is used in the loss computation for both classification and regression.}
\label{fig:pipeline}
\end{figure*}

We illustrate our framework in Fig.~\ref{fig:pipeline} and introduce the detailed implementations in the following sections. Overall, our key insight is to design a  dynamic and accurate training target for DETR. For the first concern, we build a strong connection between the classification and regression by adopting a regression-aware classification loss.  To mitigate the second issue , we adopt many-to-one matching along with a ranking \& weighting strategy. In this way, both of the two misalignment issues can be solved jointly.

\subsection{Align-DETR} \label{sec:align_loss}

% The original loss ignores the misalignment between the classification score and regression accuracy and thus is incapable of  ensuring that the BR samples have a high overlap with HC samples. 

% \textbf{Align Loss}.

Driven by the aforementioned concerns, our objective is to enhance the optimization of DETR by addressing the misalignment issue. Initially, we present our matching strategy, followed by the introduction of our proposed loss function, denoted as the Align Loss. 
This sequential approach is aimed at systematically mitigating misalignment and thereby improving the overall efficacy of DETR optimization.

\textbf{Mixed Matching and Ranking Strategy}.
DETR~\cite{detr} and most of its variants~\cite{deformabledetr,conddetr}  adopt  Hungarian  Matching to learn a unique association between GT and predictions. 
However, this approach assigns only one positive sample for each GT annotation, rendering it susceptible to the instability inherent in matching, as noted in previous works~\cite{dino,liu2023stable}.
To address this challenge, we propose a gradual transition from positive to negative samples which involves implementing a mixed-matching and ranking strategy.
% inspired by the hybrid layer matching technique noted in \cite{hdetr}, which combines both many-to-one and one-to-one matching schemes.

Given predictions $P$ and ground truth $G$, each comprising $N$ instances, we employ a modified  version of Hungarian Matching to assign $k$ predictions to each ground truth, resulting in a total of $kN$ matched samples, termed candidates. These candidates are subsequently arranged based on their distances from the GT.   We propose defining a quality metric $q$ as inspired  previous studies ~\cite{tood,autoassign}, which represents the geometric average of classification accuracy ($p$) and regression precision ($u$):
\begin{equation}
q = p^\alpha \cdot u^{(1-\alpha)},
\end{equation}
where $p$ denotes the binary classification score, $u$ signifies the IoU between the predicted bounding box and the ground truth, and $\alpha$ serves as a hyper-parameter to balance these factors. 
% The metric $q$ plays a pivotal role in guiding the model towards establishing a robust relationship between classification and regression. 
% While the conventional cost function used in matching is derived from the sum of classification and regression costs, leading to a loose coupling between classification and regression and thus is not the optimal choice.
We denote the ranking of each candidate as $r\in \{0,1,2,3, k-1\}$.  We set $k>1$ for intermediate predictions and expect the change of matching happens within a candidate bag. As for the last decoder layer, we set $k=1$ for one-to-one association.

Align-DETR shares some similarities with $\mathcal{H}$-DETR~\cite{hdetr} but diverges in both motivation and implementation: \textbf{(a)} While $\mathcal{H}$-DETR utilizes many-to-one matching primarily to expedite convergence, we employ it to ensure consistent optimization across layers; \textbf{(b)} $\mathcal{H}$-DETR treats all positive samples equally, whereas we introduce an adaptive target mechanism, as detailed in Section~\ref{sec:align_loss}.
% Thus, it builds a weaker correlation between classification score and regression accuracy.

% design a dynamic loss that helps align the classification and regression, which we term as Align Loss.
\textbf{Align Loss}.
To promote more consistent and efficient optimization, we outline two guiding principles to inform the loss design of DETR. Firstly, the target of the classification loss should be adaptive and position-guided, echoing findings in prior literature~\cite{liu2023stable}. Secondly, there should be a smooth transition from positive samples to negative samples. 

In accordance with these principles, we propose a straightforward yet effective loss function for DETR, defined as follows:
\begin{equation}
\label{eq:align_loss}
    {\cal L}_{align} = -t_c\cdot (1-p)^\gamma \cdot logp - (1-t_c) \cdot p^\gamma \cdot log(1-p),
\end{equation}
wherein the hard label $y$ in Eq.~\ref{eq:focal_loss} is substituted with a soft target $t_c$.  
As shown in Eq.~\ref{eq:align_loss},  adjusting  $t_c$ from 1 to 0 makes a smooth transition from a positive target to a negative target.
This property makes it perfectly compatible to achieve a transition between positive sample and negative sample. 
% Though such form does not guarantee a optimal point of $p'=t_c$, it still ensures a  
We define $t_c$ as follows :
 \begin{align}
    &t_c = e^{-r/\tau}\cdot q,
    \label{eq:quality}
\end{align}
where  $e^{-r/t}$ $\tau$  is an exponential down-weighting term controlled by a  hyper-parameter $\tau$. By associating the $t_c$ with the joint quality, Align Loss can guide the learning of classification with regression precision simultaneously and thus build a strong connection between these two tasks~\cite{tood,freeanchor}. In the literature, \cite{liu2023stable} employs a position-supervised classification loss to establish a unified optimization framework, which bears similarities to our approach. However, our method approaches the problem from a distinct perspective, emphasizing the alignment of the two tasks. Thus, we utilize the classification confidence in $t_c$, which our experiments in Section~\ref{sec:exp} have validated as crucial. Another noteworthy distinction lies in our identification of misalignment in the classification target across layers due to the unstable matching phenomenon. This issue is addressed in our method through a gradual positive-to-negative transition.

% Align Loss degenerates to Focal loss when $t_c$ is extremely large or small. This property makes it compatible with Foca

Given that the Align Loss functions as a "soft" variant of focal loss~\cite{focalloss}, seamlessly integrating with any DETR-variant compatible with focal loss is feasible. To leverage this capability, we propose an asymmetric classification loss by applying Align Loss on selected candidates and focal loss on background samples:
\begin{equation}
\label{eq:cls_loss}
% p_i, g_{\sigma(i)
{\cal L}_{cls} = \sum_i^{N_{pos}}  {\cal L}_{align}(p_i, y_i) + \sum_j^{N_{neg}} {\cal L}_{focal}(p_j,0),
\end{equation}
where $N_{pos}$ and $N_{neg}$ denote the number of total positive samples and negative samples, respectively. 
% For consistency, we use the same value of $\gamma=2.0$ for both losses. 

% As outlined in Stable-DINO \cite{liu2023stable}, position-supervised classification loss can also contributes to constructing a singular optimization path, thereby facilitating faster convergence. Different from Stable-DINO, we approach the optimization problem of DETR through a different view, \emph{i.e.} misalignment, and thus our method differs from it in several points: (i). Align Loss is guided by both classification score and regression precision to ensure a strong connection between these two tasks; (ii). We use many-to-one matching and a soft transition to overcome the misalignment in training signals across different layers. In Sec.~\ref{sec:exp}, comparative experiments demonstrate though sharing a similar form, Align Loss is better than the method proposed by Stable-DINO.

In the context of regression tasks, though not obviously influenced by the misalignment issue aforementioned, we opt to implement a  regression loss consistent with Align Loss. This helps achieves a consistent optimization in both tasks. 
% eliminating the $q$ term from $t_c$ and apply only the exponential down-weighting term $e^{(-r/\tau)}$ on L1 loss and GIoU loss.
Given predicted bounding box $b_i$ and GT box $\hat{b}_i$ , our regression loss is defined as follows:
\begin{equation}
\label{eq:loss_reg}
{\cal L}_{reg} = \sum_i^{N_{pos}}  e^{(-r_i/\tau)} \cdot( {\cal L}_{l1}(b_i, \hat{b}_{i}) +  {\cal L}_{GIoU}(b_i, \hat{b}_{i}))
\end{equation}
Ultimately, our loss is defined as:

\begin{equation}
\label{eq:l_reg}    
{\cal L} = \sum_{l=1}^{L-1} {\cal L}_{task} (P_l, G^{(k)}) + {\cal L}_{task} (P, G),
\end{equation}
where ${\cal L}_{task}$ is a weighted combination of classification loss ${\cal L}_{cls}$ and ${\cal L}_{reg}$ ~\cite{detr}, $G^{(k)}$ is an augmented version of GT by copying $k$ times and  $L$ is the total number of decoder layers.

In summary, the Align-DETR introduces an Align Loss along with a matching strategy to solve the misalignment issue for higher precision on localization of DETR. Without loss of generality, our method can be integrated into any DETR-like architecture.

\section{Experiments}\label{sec:exp}

\subsection{Setup}

\textbf{Datasets.} We conduct all our experiments on MS-COCO 2017~\cite{coco} Detection Track and report our results with the mean average precision  metric on the validation dataset.

\textbf{Implementation details.} 
We use  DINO~\cite{dino} as the baseline method, along with their default hyper-parameter settings.
% The DAB-DETR baseline employs a standard transformer architecture that takes single-scale features as inputs. 
The DINO baseline adopts deformable-transformer~\cite{deformabledetr} and multi-scale features as inputs. 
For the hyper-parameters introduced in Align-DETR, we set $k=4$, $\alpha=0.25$, and $\tau=1.5$. 
To ensure a fair comparison with recent methods~\cite{hu2024dac,dino, liu2023stable}, we train  \methodname for 1$\times$ and 2$\times$ schedules.
We implement our methods with the help of open-source library detrex~\cite{detrex}. To optimize the model, we set the initial learning rate to $1 \times 10^{-4}$ and decay it by multiplying $0.1$ for backbone learning. We use AdamW~\cite{loshchilov2018decoupled} as the optimizer with $1 \times 10^{-4}$ weight decay  and set  batch size to 16 for all our experiments. 
% For ablation studies, the model is trained for 12 epochs (1x schedule), and its learning rate is dropped by multiplying 0.1 at the 11th epoch. 
% For ablation study, the model is trained for 30 epochs and the learning rate is dropped by multiplying  0.1 at 25th epoch. 

% As for regression, since these two tasks should be synchronized to make predictions accurate both in classification and regression, we apply the same unique weighting factor to regression loss as: 

\begin{table}[ht]
    \centering
    \scalebox{0.9}{
    \begin{tabular}{l|c|c|ccccccccc}
        % \toprule
        \hline
        Model & \#epochs & Backbone  &AP \quad & AP$_{50}$ & AP$_{75}$ & AP$_{S}$ & AP$_{M}$ & AP$_{L}$ \\
        \hline
        % Faster RCNN~\cite{faster}  &\ $108$ & $--$ & $42.0$  & $62.1$ & $45.5$ & $26.6$ & $45.5$ & $53.4$ & $42$M & 180\\
        % SAM-DETR \cite{zhang2022accelerating} & 50 & R50 & 39.8 & 61.8 & 41.6 & 20.5 & 43.4 & 59.6 \\
        SMCA-DETR~\cite{smca}   &  $50$ & R50 &  $43.7$ &  $63.6$ &  $47.2$ &  $24.2$ &  $47.0$ &  $60.4$\\
        SAM-DETR~\cite{zhang2022accelerating} & 50 & R50 & 45.0 & 65.4 &47.9 & 26.2 & 49.0 & 63.3 \\
        Def.DETR~\cite{deformabledetr}& $50$& R50 & $45.4$ & $64.7$ & $49.0$ & $26.8$ & $48.3$ & $61.7$\\
        AdaMixer\cite{adamixer} & 36& R50 & 47.0 & 66.0 & 51.1 & 30.1 & 50.2 & 61.8\\
        SD-DETR~\cite{zhang2023decoupled} & 50 & R50 & 45.5 & 65.4 & 48.5 & 25.6 & 49.9 & 64.2\\
        DAB-Def.DETR~\cite{dabdetr} & $50$& R50 & $46.9$ & $66.0$ & $50.8$ & $30.1$ & $50.4$ & $62.5$\\
        DN-Def.DETR~\cite{dndetr} & $12$ & R50&  $43.4$ & $61.9$ & $47.2$ & $24.8$ & $46.8$ & $59.4$\\
        DN-Def.DETR~\cite{dndetr} & $50$& R50  & $48.6$ & $67.4$ & $52.7$ & $31.0$ & $52.0$ & $63.7$ \\
        DINO~\cite{dino} &  $12$& R50 & 49.0 & $66.6$ & $53.5$ & $32.0$ & $52.3$ & $63$\\
        DINO~\cite{dino} &  $24$ & R50& 50.4 & $68.3$ & $54.8$ & $33.3$ & $53.7$ & $64.8$\\

        Co-DETR~\cite{zong2023detrs} & 12& R50 & 49.5  & 67.6 & 54.3 & 32.4 &  52.7 & 63.7\\
        Cascade-DETR~\cite{ye2023cascade} & 12 & R50 & 49.7 & 67.1 & 54.1 & 32.4 & 53.5 & 65.1 \\
        Group-DETR~\cite{chen2023group}  & $12$ & R50 & $49.8$ &$--$& $--$ &$32.4$ & $53.0$ & $64.2$\\ 
        $\mathcal{H}$-DETR~\cite{hdetr} & $12$ & R50 &$48.7$ & $66.4$ & $52.9$ & $31.2$ & $51.5$ & $63.5$ \\
        DAC-DETR~\cite{hu2024dac} & $12$ & R50 &  $50.0$ & {67.6} & 54.7 & 32.9 & 53.1 & 64.2\\
        DAC-DETR~\cite{hu2024dac} & $24$ & R50 &  $51.2$ & $68.9$ & 56.0 & 34.0 & 54.6 & 65.4\\
        Salience-DETR~\cite{Hou_2024_CVPR} & 12 & R50 & 50.0 & 67.7 & 54.2 & 33.3 & 54.4 &64.4\\
        Salience-DETR~\cite{Hou_2024_CVPR} & 24 & R50 & 51.2 & 68.9 & 55.7 & 33.9 & 55.5 &65.6\\
        Rank-DETR~\cite{pu2023rank} & 12 & R50 &  50.2 & 67.7. & 55.0 & 34.1 & 53.6 & 64.0 \\
        % Rank-DETR~\cite{pu2023rank} & 36 & R50 &  50.2 & 67.7. & 55.0 & 34.1 & 53.6 & 64.0 \\
        MS-DETR~\cite{zhao2024ms} & 12 &R50 & 50.0 & 67.3 & 54.4 & 31.6 & 53.2  & 64.0\\
        MS-DETR~\cite{zhao2024ms} & 24 & R50 & 50.9 & 68.4 & 56.1 & 34.7 &54.3 & 65.1\\
        
          Focus-DETR~\cite{focusdetr} & 36 & R50  & 50.4 & 68.5 & 55.0 &{34.0} & 53.5 & 64.4\\ 
        Stable-DINO~\cite{liu2023stable} & 12 & R50 & {50.4}& 67.4& {55.0} & 32.9 & \textbf{54.0} & \textbf{65.5} \\
        Stable-DINO~\cite{liu2023stable} & 24& R50  &  51.5 & 68.5 & \underline{56.3} & 35.2 & 54.7 & \underline{66.5}\\
        % Rank-DETR~\cite{ran}
        \hline
        % \methodname & $12$ & $47.3$ & $64.7$ & $51.7$ & $29.8$ & $50.7$ & $62.8$ & $42$M & 195\\
        % \methodname-prog & $12$ & $49.2$ & $66.2$ & $53.9$ & $32.7$ & $52.8$ & $63.6$ & $47$M &279\\
        \rowcolor{gray!19}
        \methodname (Ours)& $12$ & R50& \textbf{50.5} &	\textbf{67.7} & \textbf{55.3}	 &\textbf{34.7} & {53.6} &	{64.6}\\
       \rowcolor{gray!19}
        \methodname (Ours)& $24$ & R50& \underline{51.7} & \underline{69.0} & \underline{56.3}& \underline{35.5} & \underline{55.0} & 66.1\\

        % \methodname-R$50$ & $12$ & $47.3$ & $64.7$ & $51.7$ & $29.8$ & $50.7$ & $62.8$
        % \bottomrule
        \hline
    \end{tabular}
    }
    \centering
    \caption{Comparisons (\%) of  \methodname and other DETR-like methods on COCO \emph{val} set. Def.DETR is the abbreviation of Deformable DETR. \textbf{Bold} and \underline{underlined} text are best results under 1$\times$ and 2$\times$ schedule setting, respectively. }
    \label{tab:multiscale_results}
\end{table}

\begin{table}[htb]
    \centering
    \scalebox{1.0}{
    \begin{tabular}{l|c|ccc}
    \toprule
         Method & w/ Align Loss &AP & \AP{50} & \AP{75}\\
        \hline
        $\mathcal{H}$-DETR~\cite{hdetr} & & 48.7 & 66.4 & 52.9 \\
        Align-$\mathcal{H}$-DETR & \checkmark & \textbf{49.3}&  \textbf{67.2} & \textbf{53.7}\\
        % DAC-DETR~\cite{hu2024dac}  &   \underline{67.2} & \underline{53.7} \\
        % Align-DAC-DETR~\cite{hu2024dac} & 
    \bottomrule
    \end{tabular}
    }
    \caption{Comparisons (\%) of Align-$\mathcal{H}$-DETR and $\mathcal{H}$-DETR on COCO \emph{val} set with 1x schedule.}
    \label{tab:h_detr}
\end{table}
\subsection{Main Results}

% We use \methodname~to denote our method with a plain setting  and \methodname-prog to denote our method with a progressive setting as described in Sec.~\ref{sec:eqal}.  

We conduct experiments using DINO~\cite{dino} and $\mathcal{H}$-DETR~\cite{hdetr} as the baselines, which adopts the deformable-transformer as the backbone.  DINO uses tricks such as CDN, look forward-twice, and bounding box refinement for better performance. We follow DINO's approach and adopt its tricks. Regarding to backbone, we use an Resnet-50 (R-50)~\cite{he2016deep} backbone with 4-scale features (P3, P4, P5, and P6) as input.

\begin{table}[tb]
    \centering
    \scalebox{1.0}{
    \begin{tabular}{lcccc}
        \toprule
        Method & AP & \AP{50} & \AP{75} \\    
        \midrule
        Focal Loss~\cite{focalloss} & 49.0 & 66.0 &53.5\\
        \midrule
         IoU branch  ~\cite{iounet}& 49.2 & 66.3 &53.5\\
         QFL~\cite{gfl} & 47.6&64.3 &51.8 & \\
         VFL~\cite{vfl} & 48.7& 67.0&52.3 &\\
         PSL~\cite{liu2023stable} & 49.8 & 66.7 & 54.5\\
         PSL + PMC~\cite{liu2023stable} & 50.2 & 66.7 & 55.0\\
         \rowcolor{gray!25}
         Align Loss (Ours) &\textbf{50.5}& \textbf{67.8}& \textbf{55.3}&\\
        \bottomrule
    \end{tabular}}
    \caption{Comparison (\%) with other methods on the misalignment problem on COCO \emph{val}. We use ''PSL'' and ''PMC'' for position-supervised loss and position-modulated matching in Stable-DINO~\cite{liu2023stable}}
    \label{tab:loss_comparison}
\end{table}

\begin{table}[htb]
  \centering
  \scalebox{0.9}{
    \begin{tabular}{ccc|ccc}
        \toprule
        Cls Loss & Reg Loss &  Matching & AP & \AP{50} & \AP{75}\\    
        \midrule
         \ding{51}  & \ding {51}  & \ding{51}&\textbf{50.5} & \textbf{67.8} &\textbf{55.3} \\
       \ding{51} & \ding {51}  & & 50.1 & 67.2 & 54.8\\ 
        \ding{51} &    &\ding{51}& 49.7 & 66.9 & 54.1\\
       
          & \ding {51}  &\ding{51}&  49.1 & 67.5 &  53.4\\
          &   & \ding{51}&  49.0 & 66.0 & 53.5  \\

        \bottomrule
    \end{tabular}
    }
  \caption{Ablation study (\%) of \methodname~on each component in terms of AP on COCO \emph{val}. The results demonstrate the effectiveness of our proposed component.}
  \label{tab:ab_1}
\end{table}

% We conduct three multi-scale experiments with the 1x schedule.  
% \emph{First}, we test a simple setting with plain matching, 300 queries, and no two-stage option, and it is denoted as \methodname. \emph{Second}, we strengthen the first setting with progressive matching, 900 queries, and two-stage enabled, and it is denoted as \methodname-prog. In \methodname-prog , we modify the one-to-one matching in the first stage with many-to-one matching, and the size of the candidate bag is set to $7$. \emph{Third}, to verify whether there is an overlap between \methodname~and extra branch-based methods like DINO\cite{dino}, we extend \methodname-prog with denoising mechanism  and denote it \methodname-dino.  E-QAL is applied to all predictions in our methods.

The results are presented in  Tab.~\ref{tab:multiscale_results} and Tab.~\ref{tab:h_detr}.  Despite the highly optimized structure of DINO~\cite{dino},  our method  still outperforms it by  1.5\% and 1.3\% AP  in  1$\times$ and 2$\times$ schedules, respectively.  This indicates that even the advanced DETR-variant can be affected by the misalignment problem.  Then we compare \methodname~to two recent state-of-the-art methods, DAC-DETR~\cite{hu2024dac} and Stable-DINO~\cite{liu2023stable}, and find that \methodname~achieves higher AP while using fewer tricks such as memory fusion, demonstrating its superior effectiveness.  
It's noteworthy that \methodname exhibits highly competitive performance, particularly in detecting small objects. In this domain, it surpasses Stable-DINO by 1.8\% in terms of AP, indicating that small objects are more susceptible to the misalignment issue. This highlights the effectiveness of \methodname in addressing the challenges posed by misalignment, particularly in scenarios where precise localization is crucial, such as detecting small objects.
\methodname also  outperforms other competitors such as SMCA~\cite{smca}, Faster RCNN-FPN~\cite{faster}, Deformable-DETR~\cite{deformabledetr} and Focus-DETR~\cite{focusdetr} with much less training schedule. At last, we compare our methods to two DETR-variants that also focus on improving the assignment of DETR, \emph{i.e.}  $\mathcal{H}$-DETR~\cite{hdetr} and Group-DETR~\cite{chen2023group}, and find \methodname~leads them by a large margin of 1.8\% AP and 0.7\% AP, respectively, with fewer queries used in training.  These results suggest that \methodname~is a highly effective and efficient method for object detection tasks. 

% To further validate the effectiveness of Align Loss, we also apply our Align Loss to another strong detector, $\mathcal{H}$-DETR and achieve 0.6\% AP improvement as shown in Tab.~\ref{tab:} 

\begin{table*}[htb]
  \centering
  % \begin{subtable}[t]{0.3\textwidth}
  \subfigure{
      \scalebox{0.7}{
    \begin{tabular}{c|cccc}
        \toprule
        $\alpha$ &  0  & \cellcolor{gray!15}{0.25} & 0.5 & 0.75\\    
        \midrule
        AP &50.0 &  \cellcolor{gray!15}{\textbf{50.5}} & 49.2 & 47.6\\
        \bottomrule
    \end{tabular}}
    }
    % \end{subtable}
    % \hspace{1em}
    \subfigure{ 
      \scalebox{0.7}{
    \begin{tabular}{c|ccccc}
        \toprule
        $k$ &  1  & 2 & 3 &  \cellcolor{gray!15}{4} & 5\\    
        \midrule
        AP &50.1 & 50.2 & 50.4 & \cellcolor{gray!15}{ \textbf{50.5}} & 50.2\\
        \bottomrule
    \end{tabular}
    }}
    % \hspace{1em}
    \subfigure{ 
      \scalebox{0.7}{
    \begin{tabular}{c|cccc}
        \toprule
        $\tau$ &  \cellcolor{gray!15}{1.5}  & 3 & 6 & 9\\    
        \midrule
        AP &  \cellcolor{gray!15}{\textbf{50.5}} & 50.1 & 50.0 & 49.7\\
        \bottomrule
    \end{tabular}
    }}
  \caption{Influence (\%) of hyper-paramters $\alpha$, $k$ and $\tau$ on our approach on COCO \emph{val}. }
  \label{tab:ab_hyper}
\end{table*}

\subsubsection{Comparison with Related Methods}
% \subsubsection{Comparison with Related Methods on the misalignment problem}
In addition to comparison with state-of-the-art DETR-variants, we also implement methods like Quality Focal Loss (QFL)~\cite{gfl}, and Varifocal Loss (VFL)~\cite{vfl}, Position-Supervised Loss (PSL)~\cite{liu2023stable} on DINO~\cite{dino}, and the results are presented in Tab.~\ref{tab:loss_comparison}. 
Interestingly, we find that the the IoU-branch~\cite{iounet}, which is a widely adopted component in CNN-based detectors~\cite{ge2021yolox}, brings limited improvement to the performance.
QFL~\cite{gfl}  and VFL~\cite{vfl} also perform poorly in our experiments, which suggests that they are not designed for end-to-end detectors.  
Compared to the most closely related method, PSL~\cite{liu2023stable}, Align Loss demonstrated a significant improvement of 0.7\% AP.  Even when augmented with PMC, PSL still falls short of matching the performance of Align Loss. We attribute this discrepancy to PSL's focus on optimizing paths individually for each layer, without addressing the issue of misalignment across layers. This is likely a contributing factor to the superior performance of our method.
% We speculate that this may be due to the fact that the  predictions of DETR is already classified into positive and negative. 
% And when $\gamma=0$, it is a similar form to our method. The VFL~

\subsection{Ablation Study}
We conduct a series of ablation studies with DINO baseline to validate the effectiveness of the components. All experiments here use an R50 backbone and a schedule of standard 1x training schedule.

Firstly, we validate the effectiveness of the proposed loss design and the results are summarized in Tab.~\ref{tab:ab_1}.  It is observed that both classification loss and regression loss contribute to the final performance, with the primary contribution stemming from the classification loss, as anticipated. Notably, when the regression loss is deprecated, the performance experiences a 0.8\% AP drop, underscoring the importance of consistency in Eq.\ref{eq:l_reg} in the loss design.
To further investigate the impact of the hyper-parameters we introduced,  \emph{i.e.} $\alpha$, $k$ and $\tau$, we conduct sensitivity analysis by changing one variable and keeping other variables controlled.  Our default values are $k=4$, $\alpha=0.25$, and $\tau=1.5$. As shown in Tab.~\ref{tab:ab_hyper}, $\alpha$ is has the greatest influence on the performance while $\tau$ and $k$ have moderate effects. This sensitivity analysis supports our hypothesis that $\alpha$ should be kept small to prevent effective training signals from suppression.

\section{Conclusion}

This paper investigates the optimization of DETR and identifies two aspects of the misalignment issue that could impede performance. To address these challenges, we propose a unified and straightforward solution named Align-DETR, comprising a many-to-one matching strategy and a novel loss function, referred to as Align Loss.
To mitigate the side effects of misaligned targets across layers, our matching strategy expands the number of samples assigned to a ground truth, which we term as candidates. We anticipate the matching changes to occur within a group of candidates.
The Align Loss is designed as a "soft" variant of focal loss, employing a quality metric to guide the learning of classification with respect to position. Additionally, we implement a gradual transition from positive to negative samples within a group of candidates to smooth the conflict caused by matching change. 
Competitive experimental results are achieved on the common COCO benchmark, demonstrating the superiority of Align-DETR in terms of effectiveness.

\section*
{Acknowledgements}
{This work is partly supported by the National Natural Science Foundation of China (No. 62022011), the Research Program of State Key Laboratory of Complex and Critical Software Environment, and the Fundamental Research Funds for the Central Universities.}
\bibliography{egbib}
\end{document}